\documentstyle[fullpage,11pt]{article}
\newtheorem{theorem}{Theorem}
\newtheorem{lemma}[theorem]{Lemma}

\newtheorem{corollary}[theorem]{Corollary}

\newtheorem{definition}{Definition}

\newcommand{\ital}[1]{{\/\em #1\/}}
\newcommand{\word}[1]{\mbox{\rm #1}}

\newcommand{\andd}{\wedge}
\newcommand{\Ftwo}{\word{\bf F}_2}
\newcommand{\inv}[1]{\frac{1}{#1}}
\newcommand{\Fpow}[1]{\Ftwo^{#1}}
\newcommand{\onehalf}{{\textstyle \frac{1}{2}}}
\newcommand{\pr}[2]{\word{Pr}_{#1}\left[#2\right]}
\newcommand{\booln}{\{0,1\}^n}
\newcommand{\bool}{\{0,1\}}

\newcommand{\xx}{x}
\newcommand{\comment}[1]{}
\newcommand{\cc}{c}
\newcommand{\DD}{{\cal D}}
\newcommand{\qq}{Q}

\begin{document}
\bibliographystyle{abbrv}
\title{Noise-Tolerant Learning, the Parity Problem,\\and the
Statistical Query Model}
\author{Avrim Blum \and Adam Kalai \and Hal Wasserman}
\date{School of Computer Science\\Carnegie Mellon Univeristy}
\date{\today}
\maketitle

\begin{abstract}
We describe a slightly sub-exponential time algorithm for learning
parity functions in the presence of random classification noise.
This results in a polynomial-time algorithm for the case
of parity functions that depend on only the first $O(\log n \log\log
n)$ bits of input.  This is the first known instance of an efficient
noise-tolerant algorithm for a concept class that is provably
not learnable in the Statistical Query model of Kearns
\cite{Kearns93}.  Thus, we demonstrate that the set of problems
learnable in the statistical query model is a strict subset of those
problems learnable in the presence of noise in the PAC model.

In coding-theory terms, what we give is a poly$(n)$-time algorithm for
decoding linear $k\times n$ codes in the presence of random noise for
the case of $k = c\log n 
\log\log n$ for some $c > 0$.  (The case of $k = O(\log n)$ is trivial
since one can just individually check each of the $2^k$ possible
messages and choose the one that yields the closest codeword.)

A natural extension of the statistical query model is to allow queries
about statistical properties that involve $t$-tuples of examples (as
opposed to single examples).  The second result of this paper is
to show that any class of functions learnable (strongly or weakly)
with $t$-wise queries for $t = O(\log n)$ is also weakly learnable
with standard unary queries.  Hence this natural
extension to the statistical query model does not increase the set of
weakly learnable functions.
\end{abstract}

\section{Introduction}
An important question in the study of machine learning is:
``What kinds of functions can be learned efficiently from
noisy, imperfect data?''  The statistical query (SQ) framework of
Kearns \cite{Kearns93} was designed as a useful, elegant model for
addressing this issue.
The SQ model provides a restricted interface between a
learning algorithm and its data, and has the property that any
algorithm for learning in the SQ model can automatically be converted
to an algorithm for learning in the presence of \ital{random
classification noise} in the standard PAC model.  (This result has
been extended to more general forms of noise as well
\cite{Decatur93,Decatur96}.)  The importance of the Statistical Query model is attested to by the fact
that before its introduction, there were only a few provably
noise-tolerant learning algorithms, whereas now it is recognized that
a large number of 
learning algorithms can be formulated as SQ algorithms, and 
hence can be made noise-tolerant.

The importance of the SQ model has led to the open question of whether
examples exist of problems learnable with random classification noise
in the PAC model but not learnable by statistical queries.  This is
especially interesting because one can characterize
information-theoretically (i.e., without complexity assumptions) what
kinds of problems can be learned in the SQ model
\cite{BFJKMR94}.  For example, the class of parity functions, which
{\em can} be learned efficiently from {\em non}-noisy data in the PAC
model, provably cannot be learned efficiently in the SQ model under
the uniform distribution.  Unfortunately, there is also no known efficient
non-SQ algorithm for learning them in the presence of noise
(this is closely related to the classic coding-theory problem of
decoding random linear codes).

In this paper, we describe a polynomial-time algorithm for learning
the class of parity functions that depend on only the first $O(\log n
\log\log n)$ bits of input, in the presence of random
classification noise (of a constant noise rate).  This class
provably cannot be learned in the SQ model, and thus is the first
known example of a concept class learnable with noise but not via
statistical queries.  Our algorithm has recently been shown to have
applications to the problem of determining the shortest lattice vector 
length \cite{KS01} and to various other analyses of statistical queries
\cite{Jackson00}. 

An equivalent way of stating this result is that we are given a random
$k \times n$ boolean matrix $A$, as well as an $n$-bit vector $\tilde{y}$
produced by multiplying $A$ by an (unknown) $k$-bit
message $x$, and then corrupting each bit of the resulting
codeword $y = xA$ with probability $\eta < 1/2$.  Our goal is to
recover $y$ in time poly$(n)$.  For this problem, the case of $k =
O(\log n)$ is trivial because one could simply try each of the 
$2^k$ possible messages and output the nearest
codeword found.  Our algorithm works for  $k = c\log n\log\log n$ for some
$c > 0$.  The algorithm does not actually need $A$ to be random, so
long as the noise is random and there is no other codeword within
distance $o(n)$ from the true codeword $y$.

Our algorithm can also be viewed as a slightly sub-exponential time
algorithm for learning arbitrary parity functions in the presence of
noise.  For this problem, the brute-force algorithm
would draw $O(n)$ labeled examples, and then search through all $2^n$
parity functions to find the one of least empirical error.  (A
standard argument can be used to say that with high probability, the
correct function will have the lowest empirical error.)  In contrast,
our algorithm runs in time $2^{O(n/\log n)}$, though it also requires
$2^{O(n/\log n)}$ labeled examples.  This improvement is small but
nonetheless sufficient to achieve the desired separation result.

The second result of this paper concerns a $k$-wise version of the
Statistical Query model. In the standard version, algorithms may only
ask about statistical properties of single examples. (E.g., what is
the probability that a random example is labeled positive and has its
first bit equal to 1?)  In the $k$-wise version, algorithms may ask
about properties of $k$-tuples of examples.  (E.g., what is the
probability that two random examples have an even dot-product and have
the same label?)  Given the first result of this paper, it is natural
to ask whether allowing $k$-wise queries, for some small value of $k$,
might increase the set of SQ-learnable functions.  What we show is
that for $k=O(\log n)$, any concept class learnable
from $k$-wise queries is also (weakly) learnable from unary queries.
Thus the seeming generalization of the SQ model to allow for $O(\log n)$-wise
queries does not close the gap we have demonstrated between what is
efficiently learnable in the SQ and noisy-PAC models.  Note that this
result is the best possible with respect to $k$ because the
results of 
\cite{BFJKMR94} imply that for $k = \omega(\log n)$, there are concept
classes learnable from $k$-wise queries but not unary queries.  On the other
hand, $\omega(\log n)$-wise queries are in a sense less interesting because it
is not clear whether they can in general be simulated in the presence of noise.

\subsection{Main ideas}
The standard way to learn parity functions without noise is based on
the fact that if an example can be written as a sum (mod 2) of
previously-seen examples, then its label must be the sum (mod 2) of
those examples' labels.  So, once one has found a basis, one can
use that to deduce the label of {\em any} new example (or,
equivalently, use Gaussian elimination to produce the target function
itself).

In the presence of noise, this method breaks down.  If the original
data had noise rate $1/4$, say, then the sum of $s$ labels
has noise rate $1/2 - (1/2)^{s+1}$.  This means we can add
together only $O(\log n)$ examples if we want the resulting sum to be
correct with probability $1/2 + 1/poly(n)$.  Thus, if we want to use
this kind of approach, we need some way to write 
a new test example as a sum of only a {\em small
number} of training examples.

Let us now consider the case of parity functions that depend on only
the first $k = \log n \log\log n$ bits of input.  Equivalently, we can
think of all examples as having the remaining $n-k$ bits equal to 0.
Gaussian elimination will in this case allow us to write our test
example as a sum of $k$ training examples, which is too many.  Our
algorithm will instead write it as a sum of $k/\log k = O(\log n)$
examples, which gives us the desired noticeable bias (that can then be
amplified). 

Notice that if we have seen $poly(n)$ training examples (and, say,
each one was chosen uniformly at random), we can argue existentially
that for $k = \log n \log\log n$, one should be able to write any new
example as a sum of just 
$O(\log\log n)$ training examples, since there are $n^{O(\log \log n)}
\gg 2^k$ subsets of this size (and the subsets are pairwise
independent).  So, while our algorithm is finding a smaller subset
than Gaussian elimination, it is not doing best possible.
If one {\em could} achieve, say, a constant-factor
approximation to the problem ``given a set of vectors, find the
smallest subset that sums to a given target vector'' then this would
yield an algorithm to efficiently learn the class of parity functions
that depend on the first $k = O(\log^2 n)$ bits of input.
Equivalently, this would allow one to learn parity functions over $n$ bits
in time $2^{O(\sqrt{n})}$, compared to the $2^{O(n/\log n)}$ time of
our algorithm.

\comment{
Similarly, if $k =$ So,
if we could algorithmically {\em find} the  
smallest subset of training examples that sums to our test 
example, we would have a noticeable bias and (essentially) be done.  

Unfortunately, it seems difficult to efficiently find the smallest
subset so we cannot do quite this well.  Instead, we give a weak
approximation.  Specifically, for $k = \log n
\log\log n$, the existential argument tells us there should exist
a subset of size $O(\log \log n)$ that sums to our test example; our
algorithm will, in this case, find a subset of size $O(\log n)$.  This
is a lot larger than optimal, but still better than Gaussian
elimination (which finds $O(\log n \log \log n)$ examples) and is
sufficient for our result.
}

\section{Definitions and Preliminaries}

A \ital{concept} is a boolean function on an \ital{input space}, which
in this paper will generally be $\booln$.   A \ital{concept class} is a
set of concepts.  We will be considering the problem of learning a
target concept in the presence of \ital{random classification noise}
\cite{AngluinLa88}.  In this model, there is some fixed (known or
unknown) noise rate $\eta < 1/2$, a fixed (known or unknown)
probability distribution $\DD$ over $\booln$, and an unknown target
concept $c$.  The learning algorithm may repeatedly ``press a button''
to request a labeled example.  When it does so, it receives a pair
$(\xx, \ell)$, where $\xx$ is chosen from $\booln$ according to $\DD$
and $\ell$ is the value $c(\xx)$, but ``flipped'' with probability
$\eta$.  (I.e., $\ell = c(\xx)$ with probability $1-\eta$, and
$\ell = 1-c(\xx)$ with probability $\eta$.)  The goal of the learning
algorithm is to find an 
\ital{$\epsilon$-approximation} of $c$: that is, a hypothesis
function $h$ such that $\Pr_{\xx \leftarrow \DD}[h(\xx) = c(\xx)] \geq
1-\epsilon$.

We say that a concept class $C$ is \ital{efficiently learnable in the
presence of random classification noise} under distribution $\DD$ if
there exists an algorithm ${\cal A}$ such that for any $\epsilon>0,
\delta>0, \eta < 1/2$, and any target concept $c \in C$, the algorithm
${\cal A}$ with probability at least $1-\delta$ produces an
$\epsilon$-approximation of $c$ when given access to $\DD$-random examples
which have been labeled by $c$ and corrupted by noise of rate
$\eta$.  Furthermore, ${\cal A}$ must run in time polynomial in $n$,
$1/\epsilon$, and $1/\delta$.\footnote{Normally, one would also
require polynomial dependence on $1/(1/2 -\eta)$ --- in part because
normally this is easy to achieve (e.g., it is achieved by any
statistical query algorithm).  Our algorithms run in polynomial
time for any \ital{fixed} $\eta < 1/2$, but have a
super-polynomial dependence on $1/(1/2 - \eta)$.}

A \ital{parity function} $c$ is defined by a corresponding vector $c
\in \booln$; the parity function is then given by the rule $c(x) = x
\cdot c \!\!\! \pmod{2}$.  We say that $c$ \ital{depends on only the first
$k$ bits of input} if all nonzero components of $c$ lie in its
first $k$ bits.  So, in particular, there are $2^k$ distinct parity
functions that depend on only the first $k$ bits of input.  Parity
functions are especially interesting to consider under the uniform
distribution $\DD$, because under that distribution parity functions
are pairwise uncorrelated.

\subsection{The Statistical Query model}
The Statistical Query (SQ) model can be viewed as providing a
restricted interface between the learning algorithm and the source of
labeled examples.  In this model, the learning algorithm may only
receive information about the target concept through \ital{statistical
queries}.  A statistical query is a query about some property $\qq$
of labeled examples (e.g., that the first two bits are equal and the label is
positive), along with a tolerance parameter $\tau \in
[0,1]$.  When the algorithm asks a statistical query $(\qq,\tau)$, it
is asking for the probability that predicate $\qq$ holds true for a
random correctly-labeled example, and it receives an approximation of this
probability up to $\pm \tau$. In other words, the algorithm receives a
response $\hat{P}_{\qq} \in [P_{\qq}-\tau, P_{\qq}+\tau]$, where
$P_{\qq} = \Pr_{x \leftarrow \DD}[\qq(x,c(x))]$. 
We also require each
query $\qq$ to be polynomially evaluable (that is, given $(x,\ell)$, we
can compute $\qq(x,\ell)$ in polynomial time).

Notice that a statistical query can be simulated by drawing a large
sample of data and computing an empirical average, where the size of
the sample would be roughly $O(1/\tau^2)$ if we wanted to assure an
accuracy of $\tau$ with high probability.

A concept class $C$ is \ital{learnable from statistical queries} with 
respect to distribution $\DD$ if there is a learning algorithm ${\cal
A}$ such that for any $c \in C$ and any $\epsilon>0$, ${\cal A}$ produces an
$\epsilon$-approximation of $c$ from statistical queries; furthermore,
the running time, the number of queries asked, and the inverse of the
smallest tolerance used must be polynomial in $n$ and $1/\epsilon$.

We will also want to talk about \ital{weak learning.}  An algorithm
${\cal A}$ weakly learns a concept class $C$ if for any $c \in C$ and
for \ital{some} $\epsilon < 1/2 - 1/\word{poly}(n)$,  ${\cal A}$ produces an
$\epsilon$-approximation of $c$.  That is, an algorithm weakly learns if
it can do noticeably better than guessing.

The statistical query model is defined with respect to non-noisy data.
However, statistical queries can be simulated from data corrupted by
random classification noise \cite{Kearns93}.  Thus, any concept class
learnable from statistical queries is also PAC-learnable in the
presence of random classification noise.
There are several variants to the formulation given above that
improve the efficiency of the simulation \cite{AslamDe93,AslamDe98},
but they are all polynomially related.  

One technical point: we have
defined statistical query learnability in the ``known distribution''
setting (algorithm ${\cal A}$ knows distribution $\DD$); in the
``unknown distribution'' setting, ${\cal A}$ is allowed
to ask for random unlabeled examples from the distribution $\DD$\@.
This prevents certain trivial exclusions from what is learnable from
statistical queries.

\subsection{An information-theoretic characterization}\label{sec:info}

BFJKMR \cite{BFJKMR94} prove that any concept class containing more than
polynomially many pairwise uncorrelated functions
cannot be learned even weakly in the statistical query model.
Specifically, they show the following.

\begin{definition} (Def.~2 of \cite{BFJKMR94})
For concept class $C$ and distribution $\DD$, the
\ital{statistical query dimension} SQ-DIM$(C,\DD)$ is the largest
number $d$ such that $C$ contains $d$ concepts $c_1, \ldots, c_d$ that
are nearly pairwise uncorrelated: specifically, for all $i\neq j$,
$$\left|\Pr_{x \leftarrow D}[c_i(x) = c_j(x)] - \Pr_{x \leftarrow
D}[c_i(x) \neq c_j(x)]\right| \leq 1/d^3.$$
\end{definition}

\begin{theorem} (Thm.~12 of \cite{BFJKMR94}) In order to learn $C$ to error
less than $1/2 - 1/d^3$ in the SQ model, where $d = $ SQ-DIM$(C,\DD)$,
either the number of queries or $1/\tau$ must be at least $\frac{1}{2}d^{1/3}$ 
\end{theorem}

Note that the class of parity functions over $\booln$ that depend on
only the first $O(\log n\log\log n)$ bits of input contains
$n^{O(\log \log n)}$ functions, all pairs of which are uncorrelated
with respect to the uniform distribution.  Thus, this class cannot be
learned (even weakly) in the SQ model with polynomially many queries
of $1/\word{poly}(n)$ tolerance.  But we will now show that there
nevertheless exists a polynomial-time PAC-algorithm for learning this
class in the presence of random classification noise.

\section{Learning Parity with Noise}
\subsection{Learning over the uniform distribution} 

For ease of notation, we use the ``length-$k$ parity problem'' to
denote the problem of learning a parity function over $\bool^k$, under
the uniform distribution, in the presence of random classification
noise of rate $\eta$.

\begin{theorem} 
\label{maintheorem}
The length-$k$ parity problem, for 
noise rate $\eta$ equal to any constant less than $1/2$, can be solved
with number of samples and total computation-time $2^{O(k/\log k)}$.
\end{theorem}

Thus, in the presence of noise we can learn parity functions over
$\{0,1\}^n$ with in time and sample size $2^{O(n/\log n)}$, and we can
learn parity functions over $\{0,1\}^n$ that only depend on the first
$k = O(\log n\log\log n)$ bits of the input in time and sample size
$poly(n)$.

We begin our proof of Theorem~\ref{maintheorem} with a simple lemma about
how noise becomes amplified when examples are added together.  For
convenience, if $x_1$ and $x_2$ are examples, we let $x_1 + x_2$
denote the vector sum mod 2; similarly, if $\ell_1$ and $\ell_2$ are
labels, we let $\ell_1+\ell_2$ denote their sum mod 2.

\begin{lemma}
\label{sumOK}
Let $(x_1, \ell_1), \ldots, (x_s, \ell_s)$ be examples labeled
by $c$ and corrupted by random noise of rate $\eta$.  Then
$\ell_1 + \cdots + \ell_s$ is the correct value of $(x_1 +
\cdots + x_s) \cdot c$ with probability $\onehalf + \onehalf(1-2\eta)^s$.
\end{lemma}

\smallskip
\noindent
{\bf Proof.}
Clearly true when $s=1$.  Now assume that the lemma is true for
$s-1$.  Then the probability that $\ell_1 + \cdots + \ell_s =
(x_1 + \cdots + x_s) \cdot c$ is 
$$(1-\eta)(\onehalf + \onehalf(1-2\eta)^{s-1}) + \eta(\onehalf -
\onehalf(1-2\eta)^{s-1}) = \onehalf + \onehalf(1-2\eta)^s.$$  
The lemma then follows by induction.

The idea for the algorithm is that by drawing many more examples than
the minimum needed to learn information-theoretically, we will be able
to write basis vectors such as $(1,0,\ldots,0)$ as the sum of a
relatively small number of training examples --- substantially smaller
than the number that would result from straightforward Gaussian
elimination.  In particular, for the length $O(\log n \log\log n)$
parity problem, we will be able to write $(1,0,\ldots,0)$ as the sum
of only $O(\log n)$ examples.  By Lemma \ref{sumOK}, this means that,
for any constant noise rate $\eta < 1/2$, the corresponding sum of
labels will be polynomially distinguishable from random.  Hence, by
repeating this process as needed to boost reliability, we may
determine the correct label for $(1,0,\ldots,0)$, which is
equivalently the first bit of the target vector $c$.  This process can
be further repeated to determine the remaining bits of $c$, allowing
us to recover the entire target concept with high probability.

To describe the algorithm for the length-$k$ parity problem, it will
be convenient to view each example as consisting of $a$
blocks, each $b$ bits long (so, $k = ab$) where $a$ and $b$ will be
chosen later.  We then introduce the following notation.

\begin{definition}
Let $V_i$ be the subspace of $\bool^{ab}$ consisting of those
vectors whose last $i$ blocks have all bits equal to zero.  An
\ital{$i$-sample} of size $s$ is a set of $s$ vectors
independently and uniformly distributed over $V_i$.
\end{definition}
The goal of our algorithm will be to use labeled examples from
$\bool^{ab}$ (these form a $0$-sample) to create an $i$-sample such
that each vector in the $i$-sample can be written as a sum of at most
$2^i$ of the original examples, for all $i=1,2,\ldots, a-1$.  We
attain this goal via the following lemma.

\begin{lemma}
\label{sampling}
Assume we are given an $i$-sample of size $s$.  We can in time
$O(s)$ construct an $(i+1)$-sample of size at least $s -
2^b$ such that each vector in the $(i+1)$-sample is written as the sum
of two vectors in the given $i$-sample.
\end{lemma}

\smallskip
\noindent
{\bf Proof.}
Let the $i$-sample be $x_1, \ldots, x_s$.  In these vectors, blocks
$a-i+1, \ldots, a$ are all zero.  Partition $x_1, \ldots, x_s$ based
on their values in block $a-i$.  This results in a partition having at
most $2^b$ classes.  From each nonempty class $p$, pick one vector
$x_{j_p}$ at random and add it to each of the other vectors in its
class; then discard $x_{j_p}$.  The result is a collection of vectors
$u_1, \ldots, u_{s'}$, where $s' \geq s - 2^b$ (since we discard at most
one vector per class).

What can we say about ${u}_1, \ldots, {u}_{s'}$?  First of all, each ${u}_j$ is
formed by summing two vectors in $V_i$ which have 
identical components throughout block $a-i$, ``zeroing out'' that
block.  Therefore, ${u}_j$ is in $V_{i+1}$.  Secondly, each
$u_j$ is formed by taking some $x_{j_p}$ and adding to it
a random vector in $V_i$, subject only to the condition that the random
vector agrees with $x_{j_p}$ on block $a-i$.  Therefore, each $u_j$ is
an independent, uniform-random member of $V_{i+1}$.  The vectors $u_1,
\ldots, u_{s'}$ thus form the desired $(i+1)$-sample.

Using this lemma, we can now prove our main theorem.

\smallskip
\noindent
{\bf Proof of Theorem \ref{maintheorem}.}
Draw $a2^b$ labeled examples.  Observe that these qualify
as a $0$-sample.  Now apply Lemma~\ref{sampling},  $a-1$ times, to
construct an $(a-1)$-sample.  This $(a-1)$-sample will have size at
least $2^b$.  Recall that the vectors in an $(a-1)$-sample are
distributed independently and uniformly at random over $V_{a-1}$, and
notice that $V_{a-1}$ contains only $2^b$ distinct vectors, one of
which is $(1,0,\ldots,0)$.  Hence there is an approximately $1-1/e$
chance that $(1,0,\ldots,0)$ appears in our $(a-1)$-sample.  If this
does not occur, we repeat the above process with new labeled examples.
Note that the expected number of repetitions is only constant.

Now, unrolling our applications of Lemma \ref{sampling}, observe that we
have written the vector $(1,0,\ldots,0)$ as the sum of $2^{a-1}$
of our labeled examples --- and we have done so without examining
their labels.  Thus the label noise is still random, and we can
apply Lemma~\ref{sumOK}.  Hence the sum of the labels gives us the
correct value of $(1,0,\ldots,0) \cdot c$ with probability
$\onehalf + \onehalf(1-2\eta)^{2^{a-1}}$.

This means that if we repeat the above process using new labeled
examples each time for poly$((\inv{1-2\eta})^{2^a}, b)$ times, we can
determine $(1,0,\ldots,0) \cdot c$ with probability of error
exponentially small in $ab$.  In other words, we can determine the
first bit of $c$ with very high probability.  And of course, by
cyclically shifting all examples, the same algorithm may be employed
to find each bit of $c$.  Thus, with high probability we can determine
$c$ using a number of examples and total computation-time $
\word{poly}((\inv{1-2\eta})^{2^a}, 2^b)$.

Plugging in $a = \frac{1}{2}\lg k$ and $b = 2k/\lg k$ yields the
desired $2^{O(k/\log k)}$ bound for constant noise rate $\eta$.

\subsection{Extension to other distributions}
While the uniform distribution is in this case the most interesting,  
we can extend our algorithm to work over any distribution.  In fact,
it is perhaps easiest to think of this extension as an online learning
algorithm that is presented with an arbitrary sequence
of examples, one at a time.  Given a new test example, the algorithm
will output either ``I don't know'', or else will give a prediction of
the label.  In the former case, the algorithm is told the correct
label, flipped with probability $\eta$.  The claim is that the
algorithm will, with high probability, be correct in all its
predictions, and furthermore will output ``I don't know'' only a
limited number of times.  In the coding-theoretic view,
this corresponds to producing a $1 - o(1)$ fraction of the
desired codeword, where the remaining entries are left blank.  This
allows us to recover the full codeword so long as no other codeword is
within relative distance $o(1)$.

The algorithm is essentially a form of Gaussian elimination, but where
each entry in the matrix is an element of the vector space $\Fpow{b}$
rather than an element of the field $\Ftwo$.  In particular, instead
of choosing a row that begins with a 1 and subtracting it from all
other such rows, what we do is choose one row for each initial $b$-bit
block observed: we then use these (at most $2^b-1$) rows to zero out
all the others.  We then move on to the next $b$-bit block.  If we
think of this as an online algorithm, then each new example seen
either gets captured as a new row in the matrix (and there are at most
$a(2^b-1)$ of them) or else it passes all the way through the matrix
and is given a prediction.  We then do this with multiple matrices and
take a majority vote to drive down the probability of error.

For concreteness, let us take the case of $n$ examples, each $k$ bits
long for $k = \frac{1}{4}\lg n (\lg\lg n - 2)$, and $\eta = 1/4$. We view each
example as consisting of $(\lg\lg n - 2)$ blocks, where each block has width
$\frac{1}{4}\lg n$.  We now create a series of matrices $M_1,
M_2, \ldots$ as follows.
Initially, the matrices are all empty.  
Given a new example, if its first block does not match the first block
of any row in $M_1$, we include it as a new row of $M_1$ (and output
``I don't know'').  If the
first block {\em does} match, then we subtract that row from it
(zeroing out the first block of our example) and consider the second
block.  Again, if the second block does not match any row in $M_1$ we
include it as a new row (and output ``I don't know''); otherwise, we subtract that row and consider
the third block and so on.  Notice that each example will either be
``captured'' into the matrix $M_1$ or else gets completely zeroed out
(i.e., written as a sum of rows of $M_1$).  In the latter case, we
have written the example as a sum of at most $2^{\lg\lg n - 2} 
= \frac{1}{4}\lg n$ previously-seen examples, and therefore the sum
of their labels is correct with probability at least $\frac{1}{2}(1 +
1/n^{1/4})$.  To amplify this probability, instead of making a
prediction we put the example into a new matrix $M_2$, and so on up to
matrix $M_{n^{2/3}}$.   If an example passes through {\em all}
matrices, we can then state that the majority vote is correct with
high probability.  Since each matrix has at most $2^{\frac{1}{4}\lg
n}(\lg\lg n - 2)$ rows, the total number of examples on which we fail
to make a prediction is at most $n^{11/12}\lg\lg n = o(n)$.

\subsection{Discussion}

Theorem~\ref{maintheorem} demonstrates that we can
solve the length-$n$ parity learning problem
in time $2^{o(n)}$.  However, it must be emphasized that we accomplish
this by using $2^{O(n/\log n)}$ labeled examples.  For the point of
view of coding theory, it would be useful to have an algorithm which takes time
$2^{o(n)}$ and number of examples $\word{poly}(n)$ or even $O(n)$.  We
do not know if this can be done.  Also of interest is the question of
whether our time-bound can be improved from $2^{O(n/\log n)}$ to, for
example, $2^{O(\sqrt{n}\,)}$.

It would also be desirable to reduce our algorithm's dependence on
$\eta$.  This dependence comes from Lemma \ref{sumOK}, with $s = 2^{a-1}$.
For instance,  consider the problem of learning parity functions
that depend on the first $k$ bits of input for $k = O(\log n\log \log
n)$. In this case, if we set $a=\lceil \frac{1}{2}\lg\lg n \rceil$ and
$b = O(\log n)$, the running time is polynomial in $n$, with
dependence on $\eta$ of $(\inv{1-2\eta})^{\sqrt{\log n}}$.  This
allows us to handle $\eta$ as large
as $1/2 - 2^{-\sqrt{\log n}}$ and still have polynomial running time.
While this can be improved slightly,
we do not know how to solve
the length-$O(\log n \log \log n)$ parity problem in polynomial time
for $\eta$ as large as $1/2 - 1/n$ or even $1/2 - 1/n^\varepsilon$.
What makes this interesting is that it is an open question (Kearns,
personal communication) whether noise tolerance can in general be
boosted; this example suggests why such a result may be
nontrivial.

\comment{
\begin{corollary}
\label{highnoise}
Let $\varepsilon$ be any positive constant.  Using $a =
\lceil\varepsilon\lg\lg n\rceil$, $b = O(\log n)$ in
Theorem~\ref{mainab}, we find that the parity problem of length
$O(\log n \log\log n)$, for noise rate $\eta \leq 1/2 - 2^{-(\lg
n)^{1-\varepsilon}}$, can be solved with number of samples and total
computation-time $\word{poly}(n)$.
\end{corollary}
}

\section{Limits of O(log n)-wise Queries}

We return to the general problem of learning a target concept $\cc$
over a space of examples with a fixed distribution $\cal D$.  A
limitation of the statistical query model is that it permits only what
may be called \ital{unary} queries.  That is, an SQ algorithm can
access $\cc$ only by requesting approximations of probabilities of
form $\pr{x}{\qq(x,\cc(x))}$, where $x$ is $\cal D$-random and $\qq$
is a polynomially evaluable predicate.  A natural question is whether
problems not learnable from such queries can be learned, for example,
from binary queries: i.e., from probabilities of form
$\pr{x_1,x_2}{\qq(x_1,x_2,\cc(x_1),\cc(x_2))}$.  The following theorem
demonstrates that this is not possible, proving that $O(\log n)$-wise
queries are no better than unary queries, at least with respect to
weak-learning.  

We assume in the discussion below that all algorithms also have access
to individual \ital{unlabeled} examples from distribution $\DD$, as is
usual in the SQ model.

\begin{theorem}
\label{lognogood}
Let $k = O(\log n)$, and assume that there exists a $\word{poly}(n)$-time
algorithm using $k$-wise statistical queries which weakly learns a concept
class $C$ under distribution $\cal D$.  That is, this algorithm learns from
approximations of $\pr{\vec{x}}{\qq(\vec{x},\cc(\vec{x}))}$, where $\qq$ is a
polynomially evaluable predicate, and $\vec{x}$ is a k-tuple of examples.
Then there exists a $\word{poly}(n)$-time algorithm which weakly learns the
same class using only unary queries, under $\cal D$.
\end{theorem}

\smallskip
\noindent
{\bf Proof.}
We are given a $k$-wise query
$\pr{\vec{x}}{\qq(\vec{x},\cc(\vec{x}))}$.  The first thing our
algorithm will do is use $Q$ to construct several candidate weak
hypotheses.  It then tests whether each of these hypotheses is in fact
noticeably correlated with the target
using unary statistical queries.  If none of them appear to be good,
it uses this fact to 
estimate the value of the $k$-wise query.  We prove that for any
$k$-wise query, with high probability we either succeed in finding a
weak hypothesis or we output a good estimate of the $k$-wise query.

For simplicity, let us assume that $\pr{x}{c(x) = 1} = 1/2$; i.e., a
random example is equally likely to be positive or negative.  (If
$\pr{x}{c(x) = 1}$ is far from $1/2$ then weak-learning is easy by
just predicting all examples are positive or all examples are
negative.)  This assumption implies that if a hypothesis $h$ satisfies
$|\pr{x}{h(x) = 1 \wedge c(x) = 1} - \frac{1}{2}\pr{x}{h(x) = 1}| \geq
\epsilon$, then either $h(x)$ or $1 - h(x)$ is a weak hypothesis.

We now generate a set of candidate hypotheses by choosing one random
$k$-tuple of 
unlabeled examples $\vec{z}$.  For each $1 \leq i \leq k$ and $\vec{\ell} \in
\{0,1\}^k$, we hypothesize 
$$h_{\vec{z},i,\vec{\ell}}(x) =
Q(z_i,\ldots,z_{i-1},x,z_i,\ldots,z_k,\vec{\ell}),$$
and then use a unary statistical query to
tell if $h_{\vec{z},i,\vec{\ell}}(x)$ or
$1-h_{\vec{z},i,\vec{\ell}}(x)$ is a weak hypothesis.  As noted above,
we will have found a weak hypothesis if 
$$
\left|\pr{x}{\qq(z_1,\ldots,z_{i-1},x,z_{i+1},\ldots,z_k,\vec{\ell})
\wedge \cc(x)=1} - 
\frac{1}{2}\pr{x}{\qq(z_1,\ldots,z_{i-1},x,z_{i+1},\ldots,z_k,\vec{\ell})}\right| \geq
\epsilon.
$$
We repeat this process for $O(1/\epsilon)$
randomly chosen $k$-tuples $\vec{z}$.  We now consider two cases.

{\bf Case I:} Suppose that the $i$th label matters to the $k$-wise
query $Q$ for some $i$ and
$\vec{\ell}$.  By this we mean there is at least an $\epsilon$ chance of the
above inequality holding for random $\vec{z}$.  Then with high probability we
will discover such a $\vec{z}$ and thus weak learn.

{\bf Case II:} Suppose, on the contrary, that for no $i$ or $\vec{\ell}$ does
the $i$th label matter, i.e.\ the probability of a random $z$
satisfying the above inequality is less than $\epsilon$.   This means
that
\begin{eqnarray*}
{\bf E}_{\vec{z}}\left[\left|\pr{x}{\qq(z_1,\ldots,z_{i-1},x,z_{i+1},\ldots,z_k,\vec{\ell})
\wedge \cc(x)=1} -  \right. \right. \\
\left. \left.
\frac{1}{2}\pr{x}{\qq(z_1,\ldots,z_{i-1},x,z_{i+1},\ldots,z_k,\vec{\ell})}\right|\right] <
2\epsilon. 
\end{eqnarray*}
By bucketing the $\vec{z}$'s according to the values of $c(z_1)$,
$\ldots$, $c(z_{i-1})$ we see that the above implies that
for all $b_1, \ldots, b_{i-1}$ $\in$ 
$\{0,1\},$
\begin{eqnarray*}
\left|\pr{\vec{z}}{\qq(\vec{z},\vec{\ell}) \wedge c(z_1)=b_1 \wedge
\ldots \wedge c(z_{i-1})=b_{i-1} \wedge c(z_i)=1}
 - \right. \\
\left. \frac{1}{2}\pr{\vec{z}}{\qq(\vec{z},\vec{\ell}) \wedge c(z_1)=b_1 \wedge \ldots \wedge
c(z_{i-1})=b_{i-1}} \right| <
2\epsilon.
\end{eqnarray*}
By a straightforward inductive argument
on $i$, we conclude that for every $\vec{b} \in \{0,1\}^k$, $$\left|\pr{\vec{z}}{\qq(\vec{z},\vec{\ell}) \wedge
c(\vec{z})=\vec{b}} -
\frac{1}{2^k}\pr{\vec{z}}{\qq(\vec{z},\vec{\ell})}\right| <
4\epsilon(1-\frac{1}{2^k}).$$ 
This fact now allows us to estimate our desired $k$-wise query
$\pr{\vec{z}}{\qq(\vec{z},\cc(\vec{z}))}$.  In particular, 
$$\pr{\vec{z}}{\qq(\vec{z},\cc(\vec{z}))} = \sum_{\vec{\ell} \in \{0,1\}^k}
\pr{\vec{z}}{\qq(\vec{z},\vec{\ell}) \wedge \cc(\vec{z})=\vec{\ell}}.$$ 
We  approximate each of the $2^k= \word{poly}(n)$ terms corresponding to a
different $\vec{\ell}$ by using {\em unlabeled} data to estimate
$\frac{1}{2^k}\pr{\vec{z}}{Q(\vec{z},{\vec{\ell}})}$.   Adding up
these terms gives us a good estimate of
$\pr{\vec{z}}{\qq(\vec{z},\cc(\vec{z}))}$ with high
probability.

\subsection{Discussion}

In the above proof, we saw that either the data is statistically
``homogeneous'' in a way which allows us to simulate the original
learning algorithm with unary queries, or else we discover a
``heterogeneous'' region which we can exploit with an alternative
learning algorithm using only unary queries.  Thus any concept class
that can be learned from $O(\log n)$-wise
queries can also be weakly learned from unary queries.  Note that Aslam and
Decatur \cite{AslamDe93} have shown that weak-learning statistical
query algorithms can be boosted to strong-learning algorithms, if they
weak-learn over \ital{every} distribution.  Thus, any concept class
which can be 
(weakly or strongly) learned from $O(\log n)$-wise queries over
\ital{every} distribution can be strongly learned over every
distribution from unary queries.

It is worth noting here that $k$-wise queries can be used to
solve the length-$k$ parity problem.  One simply asks, for each $i \in
\{1, \ldots, k\}$, the query:
``what is the probability that $k$ random examples form a basis for
$\bool^k$ and,
upon performing Gaussian elimination, yield a target concept whose
$i$th bit is equal to 1?''  Thus, $k$-wise
queries cannot be reduced to unary queries for $k = \omega(\log n)$.
On the other hand, it is not at all clear how to simulate such queries
in general from noisy examples.

\section{Conclusion}

In this paper we have addressed the classic problem of
learning parity functions in the presence of random noise.  We have
shown that parity functions over $\booln$ can be learned in slightly
sub-exponential time, but only if many labeled examples are available.
It is to be hoped that future research may reduce both the time-bound
and the number of examples required.

Our result also applies to the study of statistical query learning and
PAC-learning.  We have given the first known noise-tolerant
PAC-learning algorithm which can learn a concept class not learnable
by any SQ algorithm.  The separation we have established between the two
models is rather small: we have shown that a specific parity problem
can be PAC-learned from noisy data in time $\word{poly}(n)$, as
compared to time $n^{O(\log\log n)}$ for the best SQ algorithm.  This
separation may well prove capable of improvement and worthy of
further examination.  Perhaps more importantly, this suggests the
possibility of interesting new noise-tolerant PAC-learning algorithms
which go beyond the SQ model.

We have also examined an extension to the SQ model in terms of
allowing queries of arity $k$.  We have shown that for $k=O(\log n)$,
any concept class learnable in the SQ model with $k$-wise queries is
also (weakly) learnable with unary queries.  On the other hand, the
results of \cite{BFJKMR94} imply this is not the case for $k =
\omega(\log n)$.  An interesting open question is whether every concept
class learnable from $O(\log n \log\log n)$-wise queries is also
PAC-learnable in the presence of classification noise.  If so, then
this would be a generalization of the first result of this paper.

\newcommand{\etalchar}[1]{$^{#1}$}

\end{document}